\documentclass[conference]{IEEETran}
\usepackage{float}
\usepackage[utf8]{inputenc} 
\usepackage[T1]{fontenc}    
\usepackage{hyperref}       
\usepackage{url}            
\usepackage{booktabs}       
\usepackage{amsfonts}       
\usepackage{nicefrac}       
\usepackage{microtype}      
\usepackage{xcolor}         
\usepackage{amsmath,amssymb}
\usepackage{wrapfig}
\usepackage{stfloats}
\usepackage{graphicx}
\usepackage{caption}
\usepackage{subcaption}
\usepackage{bm}

\usepackage{listings}
\NewDocumentCommand{\codeword}{v}{%
\texttt{\textcolor{blue}{#1}}%
}

\newcommand{\Rb}{\mathbb{R}}

\title{Signature-Weighted Kolmogorov-Arnold Networks for Time Series}

\author{%
  \IEEEauthorblockN{%
    Hugo Inzirillo\IEEEauthorrefmark{1}\textsuperscript{\textsection} and
    Rémi Genet\IEEEauthorrefmark{2}\textsuperscript{\textsection}
  }%
  
  \IEEEauthorblockA{\IEEEauthorrefmark{1} CREST-ENSAE, Institut Polytechnique de Paris}%
    \IEEEauthorblockA{\IEEEauthorrefmark{2} DRM, Université Paris Dauphine - PSL}%
}

\begin{document}
\thispagestyle{plain}
\pagestyle{plain}
\maketitle
\begingroup\renewcommand\thefootnote{\textsection}
\footnotetext{These authors contributed equally.}
\endgroup

\begin{abstract}
We propose a novel approach that enhances multivariate function approximation using learnable path signatures and Kolmogorov-Arnold networks (KANs). We enhance the learning capabilities of these networks by weighting the values obtained by KANs using learnable path signatures, which capture important geometric features of paths. This combination allows for a more comprehensive and flexible representation of sequential and temporal data. We demonstrate through studies that our SigKANs with learnable path signatures perform better than conventional methods across a range of function approximation challenges. By leveraging path signatures in neural networks, this method offers intriguing opportunities to enhance performance in time series analysis and time series forecasting, among other fields.
\end{abstract}
\thispagestyle{plain}
\pagestyle{plain}

\section{Introduction}
Forecasting multivariate time series has generated interest from researchers \cite{wei2018multivariate,marcellino2006comparison}, and it has grown to be a significant field of study for many businesses. Since there is a growing amount of data available, it makes sense to suggest increasingly sophisticated prediction frameworks. Multivariate time series include multiple interdependent variables, as opposed to univariate time series, which evaluate a single variable over time. The learning task is made considerably more challenging due to these relationships. Complex models can capture temporal dependencies and dynamic interactions across several dimensions and time horizons are necessary for handling multivariate time series data. Researchers investigated a range of techniques to address the difficulties related to multivariate time series forecasting, such as neural networks \cite{tang1991time}, state-space models \cite{hamilton1994state,kim1994dynamic}, vector autoregressive models \cite{zivot2006vector,lutkepohl2013vector}, and neural networks \cite{binkowski2018autoregressive,ma2019novel}. Due to the capacity to manage nonlinearities and long-range dependencies in data, deep learning techniques like Long Short-Term Memory (LSTM) networks and attention mechanisms have demonstrated encouraging results \cite{hochreiter1997long,vaswani2017attention}. 
Recently, the Kolmogorov-Arnold Network (KAN) \cite{liu2024kan} stands out due to its theoretical foundation in the Kolmogorov-Arnold representation theorem \cite{kolmogorov1961representation} This theorem ensures that any multivariate continuous function can be decomposed into a sum of continuous one-dimensional functions, which is used by KAN to effectively approximate complex functions. Despite their theoretical robustness, there is still much room to improve the expressiveness and efficiency of KANs, especially when processing sequential and temporal data. Even more recently, some research proposed to extend KANs using wavelets \cite{bozorgasl2024wav}, or proposed new framework to incorporate KAN within RRNs \cite{genet2024tkan}. Some researchers proposed a framework to use KANs in time series analysis \cite{vaca2024kolmogorov}, and model architecture for time series forecasting using attention \cite{genet2024temporal}. When dealing with sequential and temporal data, the importance of context and feature representation is crucial for accurate and reliable predictions. These highly complex models require a "context" form of representation, summarizing the information in order to improve predictions. Path signatures, originating from rough path theory, offer a powerful method for encoding the essential features of paths or trajectories. These signatures capture the underlying geometry of the paths through a sequence of iterated integrals, making them particularly well-suited for tasks involving complex sequential data. Integrating path signatures into neural network architectures has shown promise in various applications, but their potential has not been fully explored in the context of KANs. Path signatures, first described by~\cite{chen1958integration}, are collections of iterated integrals of (transformed) time series, in this case, a series of prices for digital assets. We redirect the reader to \cite{lyons2014rough,lyons1998differential,chevyrev2016primer} for a thorough explanation of path signatures as a trustworthy representation or a set of characteristics for unparameterized paths. A path signature, also referred to as a signature for short, is essentially a mathematical expression that "succinctly" captures the structure of a path. While it originated from stochastic analysis and rough path theory, it has recently been applied to a wide range of other fields, including computer vision (\cite{yang2022developing,li2017lpsnet}), time series analysis (\cite{gyurko2013extracting,dyer2021deep}), machine learning (\cite{chevyrev2016primer,perez2018signature,fermanian2021embedding}), etc. In this paper, we introduce learnable path signatures as a novel improvement to the original KAN layer proposed by \cite{liu2024kan}. We prosposed a learnable path signature layer that uses learnable parameters to compute path signatures for each input path. The KAN framework then combines these path characteristics with conventional linear transformations. The objective of this integration is to enhance approximation skills of KANs by using the rich geometric information that path signatures provide. Codes are available at \href{https://github.com/remigenet/SigKAN}{SigKAN repository} and can be installed using the following command: \codeword{pip install sigkan}. Additionally, we release a package needed to compute the path signature using Tensorfow v2.0. \href{https://github.com/remigenet/iisignature-tensorflow-2}{iisignature-tensorflow-2} and can be installed using the following command: \codeword{pip install iisignature-tensorflow-2}. This package allows signatures to be trainable in custom tensorflow layer, although it does not support GPU utilization.

\section{Kolmogorov-Arnold Networks (KANs)}
Multi-Layer Perceptrons (MLPs) \cite{hornik1989multilayer} extension of the original percetron proposed by \cite{rosenblatt1958perceptron}, are inspired by the universal approximation theorem \cite{cybenko1989approximation} which states that a feed-forward network (FFN) with a single hidden layer containing a finite number of neurons can approximate continuous functions on compact subsets of $\Rb^n$. Kolmogorov-Arnold Network (KAN) focuses on the Kolmogorov-Arnold representation theorem \cite{kolmogorov1961representation}. The Kolmogorov-Arnold representation theorem states that any multivariate continuous function can be represented as a composition of univariate functions and the addition operation, 
\begin{equation}
    f(x_1, \dots, x_n) = \sum_{q=1}^{2n+1} \Phi_q \left( \sum_{p=1}^n \phi_{q,p}(x_p) \right),
    \label{eq:KART}
\end{equation}
where $\phi_{q,p}$ are univariate functions that map each input variable $(x_p)$ such $\phi_{q,p}: [0,1] \rightarrow \Rb$ and $\phi_q : \Rb \rightarrow \Rb$. Since all functions to be learned are univariate functions, we can parametrize each 1D function as a B-spline curve, with learnable coefficients of local B-spline basis functions. The key insight comes when we see the similarities between MLPs and KAN. In MLPs, a layer includes a linear transformation followed by nonlinear operations, and you can make the network deeper by adding more layers. Let us define what a KAN layer is
\begin{align}
    {\mathbf\Phi}=\{\phi_{q,p}\},\qquad p=1,2,\cdots,n_{\rm in},\qquad q=1,2\cdots,n_{\rm out},
\end{align}
\begin{figure}[H]
    \centering
    \includegraphics[width=0.5\linewidth]{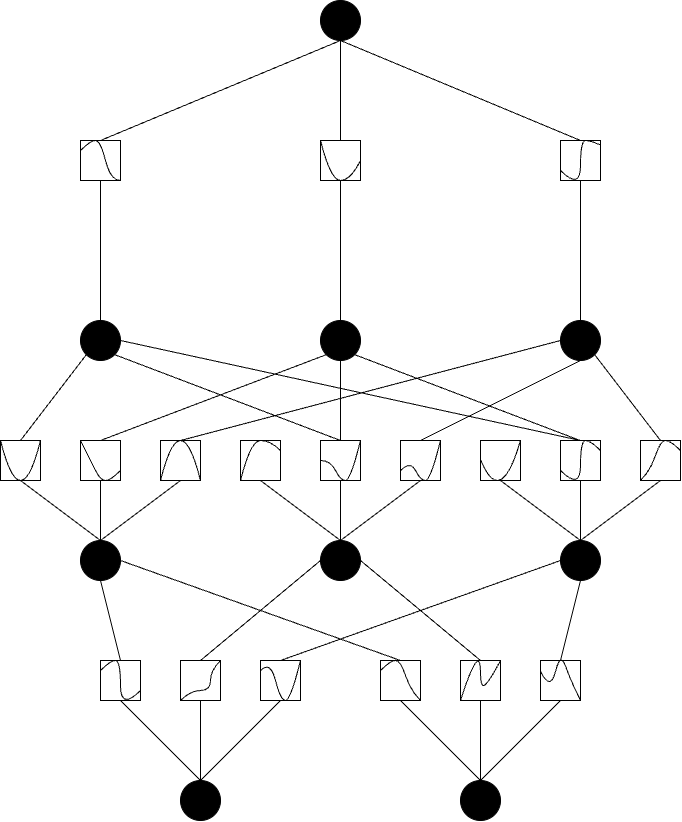}
    \caption{KAN: Kolmogorov Arnold Networks} 
    \label{fig:KAN}
\end{figure}
where $\phi_{q,p}$ are parametrized functions of learnable parameters. In the Kolmogov-Arnold theorem, the inner functions form a KAN layer with $n_{\rm in}=n$ and $n_{\rm out}=2n+1$, and the outer functions form a KAN layer with $n_{\rm in}=2n+1$ and $n_{\rm out}=1$. So the Kolmogorov-Arnold representations in Eq.~(\ref{eq:KART}) are simply compositions of two KAN layers. Now it becomes clear what it means to have Deep Kolmogorov-Arnold representation. Taking the notation from \cite{liu2024kan} let us define a shape of KAN $[n_0,n_1,\cdots,n_L],$ where $n_i$ is the number of nodes in the $i^{\rm th}$ layer of the computational graph. We denote the $i^{\rm th}$ neuron in the $l^{\rm th}$ layer by $(l,i)$, and the activation value of the $(l,i)$-neuron by $x_{l,i}$. Between layer $l$ and layer $l+1$, there are $n_ln_{l+1}$ activation functions: the activation function that connects $(l,i)$ and $(l+1,j)$ is denoted by 
\begin{align}
    \phi_{l,j,i},\quad l=0,\cdots, L-1,\quad i=1,\cdots,n_{l},\quad j=1,\cdots,n_{l+1}.
\end{align}
The pre-activation of $\phi_{l,j,i}$ is simply $x_{l,i}$; the post-activation of $\phi_{l,j,i}$ is denoted by $\tilde{x}_{l,j,i}\equiv \phi_{l,j,i}(x_{l,i})$. The activation value of the $(l+1,j)$ neuron is simply the sum of all incoming post-activations: 
\begin{equation}\label{eq:kanforward}
    x_{l+1,j} =  \sum_{i=1}^{n_l} \tilde{x}_{l,j,i} = \sum_{i=1}^{n_l}\phi_{l,j,i}(x_{l,i}), \qquad j=1,\cdots,n_{l+1}.
\end{equation}
A general KAN network is a composition of $L$ layers: given an input vector  $\mathbf{x}_0\in\mathbb{R}^{n_0}$, the output of KAN is
\begin{equation}\label{eq:KAN_forward}
    {\rm KAN}(\mathbf{x}) = (\mathbf{\Phi}_{L-1}\circ \mathbf{\Phi}_{L-2}\circ\cdots\circ\mathbf{\Phi}_{1}\circ\mathbf{\Phi}_{0})\mathbf{x}.
\end{equation}
\section{Model Architecture}
\begin{figure}[H]
    \centering
    \includegraphics[width=0.6\linewidth]{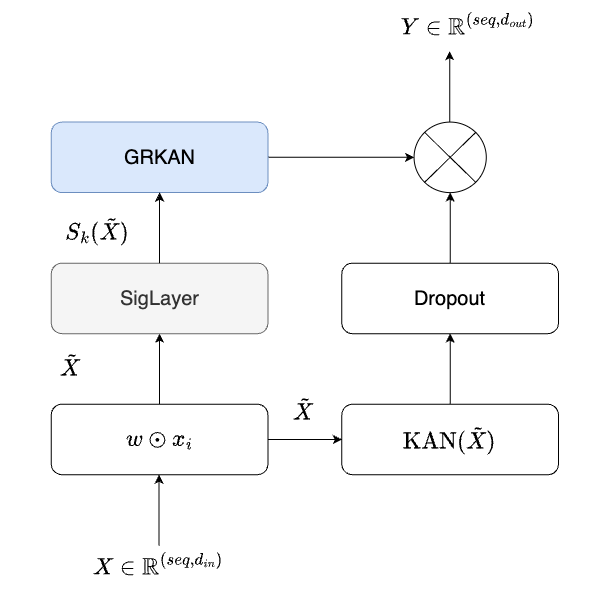}
    \caption{SigKAN} 
    \label{fig:SigKAN}
\end{figure}
In this section, we will dive into the various components of the model. We will detail each layer and mechanism. We aim to provide a comprehensive understanding of how the model operates and processes data to achieve its predictive capabilities. Before starting, we define the main components of the SigKAN:
\begin{itemize}
    \item \textbf{Gated Residual KAN:} This layer enables the modulation of information flow during the learning task.
    \item \textbf{Learnable path signature layer:} This layer will compute the path signatures for each sequence with learnable parameters.
\end{itemize}
In simple words, the Signature-Weighted KANs (SigKANs) are composed of SigLayers and KAN layers. As path signatures capture important geometric features of paths, combining the set of information with the output of KAN layers will enhance the power of prediction. Let us define a N-dimensional path $(X_t)_{t \in [ 0,T ]}$ and $X_t= (X_{1,t},...,X_{N,t})$. The  1-dimensional coordinate paths will be denoted as $ (X_{n,t})$, $n\in \{1,\ldots,N\}$.
To build the path signature of $(X_t)$, we first consider the increments of $X^1, \ldots, X^N$ over any interval $[0, t]$, $t \in [0, T]$, which are denoted $S(X)^1_{0,  t}, \ldots, S(X)^N_{0, t}$ and defined as follows:
\begin{equation}
   S(X)^n_{0, t} =  \int_0^{t} dX^n_s.
\end{equation}
$S(X)^n_{0, t}$ is the first stage to calculate the signature of an unidimensional path. It is a sequence of real numbers, each of these numbers corresponding to an iterated integral of the path. 
Then, the next signature coefficients will involve two paths: a coordinate path $(X^m_t)$ and the increment path{{}} $(S(X)^n_{0, t})$ associated to the coordinate path $(X_t^n)$. 
There are $N^2$ such second order integrals, which are denoted $S(X)^{1,1}_{0, t} \ldots, S(X)^{N,N}_{0, t}$, where
\begin{equation}
    S(X)^{n,m}_{0, t} = \int_0^t S(X)^n_{0, s} \, dX^m_s,\; n,m\in\{1,\ldots,N\}. 
\end{equation}
The set of first (resp. second) order integrals involves $N$ (resp. $N^2$) integrals. The latter first and second order integrals are called the first and second levels of path signatures respectively. Iteratively, we obtain $N^k$ integrals of order $k$, which is denoted $S(X)^{i_{1},..,i_{k}}_{0, t}$ for the $k$-th level of path signatures, when $i_j\in \{1,\ldots,N\}$ and $j\in \{1,\ldots, k\}$.
More precisely, the $k$-th level signatures can be written
$$
S(X)^{i_{1},\ldots,i_{k}}_{0, t} =  \int_0^t S(X)^{i_1, \ldots, i_{k-1}}_{0, s} \, dX^{i_k}_s.
$$
The path signature $S(X)_{0,T}$ is finally the infinite ordered set of such terms when considering all levels $k\geq 1$ and the path on the whole interval $[0,T]$: 
\begin{equation}
\begin{split}
    S(X)_{0,T} & = (1, S(X)^1_{0,T}, S(X)^2_{0,T}, \ldots, S(X)^N_{0,T}, S(X)^{1, 1}_{0,T},\\
    & S(X)^{1, 2}_{0,T}, \ldots, S(X)^{N, N}_{0,T}, S(X)^{1, 1, 1}_{0,T}, \ldots).
\end{split}
\end{equation}

\subsection{Gated Residual KANs}
The relationship between temporal data is a key issue. Gated Residual Networks (GRNs) offer an efficient and flexible way of modeling complex relationships in data. They allow controlling the flow of information and facilitate the learning tasks. They are particularly useful in areas where nonlinear interactions and long-term dependencies are crucial. In our model, we use the Gated Residual Kolmogorov-Arnold Networks (GRKAN) inspired by the GRN proposed by \cite{lim2021temporal}, we kept the same architecture. We propose a new approach using two KAN layers to control the information flow while bringing more interpretability. Using GRKAN, there is no more need for context which is contained in path signature; an additional linear layer is required to match the signature transform and the output of the gating mechanism.
\begin{figure}
    \centering
    \includegraphics[width=0.6\linewidth]{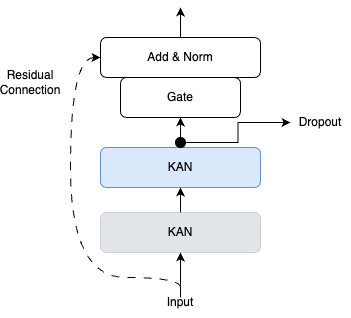}
    \caption{Gated Residual KAN (GRKAN)} 
    \label{fig:GRKAN}
\end{figure}
\begin{align}
\text{GRN}_\omega\left(x \right) &=\text{LayerNorm}\left(x + \text{GLU}_\omega(\eta_1) \right), \\
    \eta_1 &= \text{KAN}(\varphi_{\eta_1}(.),\eta_2),  \label{eqn:grn_step}\\
    \eta_2 &= \text{KAN}(\varphi_{\eta_2}(.),x). & 
\label{eqn:grkan}    
\end{align}
In this context, we used activation functions for KAN layers denoted, $\varphi_{\eta_1}(.)$ and $\varphi_{\eta_2}(.)$, $\text{SiLU}$ \cite{elfwing2018sigmoid} and ELU \cite{clevert2015fast}, respectively, while $\eta_1 \in \mathbb{R}^{d_{model}}$ and $\eta_2 \in \mathbb{R}^{d_{model}}$ represent intermediate layers. The standard layer normalization $\text{LayerNorm}$ is that described in \cite{ba2016layer}, and $\omega$ is an index used to indicate weight sharing. When the expression $\text{KAN}(x)$ is largely positive, ELU activation works as an identity function. On the other hand, when this expression is largely negative, ELU activation produces a constant output, thus behaving like a linear layer. We used Gated Linear Units (GLUs) \cite{dauphin2017language} to provide the flexibility to suppress any parts of the architecture that are not required for a given dataset. Letting $\gamma \in \mathbb{R}^{d_{model}}$ be the input, the GLU then takes the form:
\begin{align}
 \text{GLU}_\omega(\gamma) & =  \sigma(W_{4, \omega}~\gamma + b_{4, \omega}) \odot (W_{5, \omega}~\gamma + b_{5, \omega} ),
\label{eqn:component_gate}
\end{align}
where $\sigma(.)$ is the sigmoid activation function, $W_{(.)} \in \mathbb{R}^{d_{model}\times d_{model}}, b_{(.)} \in \mathbb{R}^{d_{model}}$ are the weights and biases, $\odot$ is the element-wise Hadamard product, and $d_{model}$ is the hidden state size. GLU allows to control the extent to which the GRN contributes to the original input $x$ potentially skipping over the layer entirely if necessary as the GLU outputs could be all close to 0 in order to suppress the nonlinear contribution. 

\subsection{Learnable Path Signature}
For a path \(X_n\) represented as a sequence of points \((x_{n,1}, x_{n,2}, \ldots, x_{n,N})\), we transform each point with learnable coefficients:
\begin{equation}
    \tilde{X}_{n,i} = w_n \odot x_{n,i}
\end{equation}
where $w_n$ is a learnable vector of coefficients and $\odot$ denotes element-wise multiplication. The $k$-th order signature can be expressed as:

\begin{equation}
\small
\begin{split}
S(\tilde{X}) = \left(1, \left( \int_0^1 d\tilde{X}^{i_1}_{t_1} \right)_{1 \leq i_1 \leq N}, \left( \int_0^1 \int_0^{t_1} d\tilde{X}^{i_2}_{t_2} d\tilde{X}^{i_1}_{t_1} \right)_{1 \leq i_1, i_2 \leq N}, \ldots, \right. \\
\left. \left( \int_0^1 \int_0^{t_1} \cdots \int_0^{t_{k-1}} d\tilde{X}^{i_k}_{t_k} \cdots d\tilde{X}^{i_2}_{t_2} d\tilde{X}^{i_1}_{t_1} \right)_{1 \leq i_1, i_2, \ldots, i_k \leq N}, \ldots \right).
\end{split}
\end{equation}
After performing this step, we will compute the path signature of each sequence of transformed points $(\Tilde{X}_1,...,\Tilde{X}_N)$. The transformed path \(\tilde{X}_1\) is now a vector of transformed points $(\tilde{x}_{1,1}, \tilde{x}_{1,2}, \ldots, \tilde{x}_{1,N})$. For $N$ paths $(X_1,...,X_N)$ the path signature transformation:
\begin{equation}
    S(\Tilde{X}) = \left[S(\tilde{X})_1, S(\tilde{X})_2, S(\tilde{X})_3,....\right],
    \label{eq:sig_vec}
\end{equation}
The path signature obtained provides a detailed representation of the underlying path, capturing patterns and dependencies across the data. This can be highly informative for the prediction, and potentially lead to better performance. To do so, the output from \eqref{eq:sig_vec} will be the input of a GRKAN \ref{fig:GRKAN}, it is possible to use another gated mechanism to control the flow of information, enabling the networks to focus on the most relevant features and suppress less useful ones.
\begin{equation}
    h_s =f(S_k(\Tilde{X})),
\end{equation}
where $f(.)$ is a succession of operations, in our case we choose to use a Gated Residual KAN (GRKAN), the output of this channel will weight the output of the different KAN layers.

\subsection{Output}
We denoted $seq$ and $d_{out}$, the sequence length and the output dimension, respectively. The output of the GRKAN is used as a weighting applied on the output of a KAN layer on $(\Tilde{X})$. Let us denote the output of the GRKAN as $h_s$, this one is then passed through a softmax activation
\begin{equation}
    \psi = \text{SoftMax} (h_s).
\end{equation}
The global output of the SigKAN layer is obtained from
\begin{equation}
\psi \odot \text{KAN}(\Tilde{X}).
\end{equation}
It is to note that $\Tilde{X}$ is a matrix of shape $(seq, d_{out})$, while $KAN(\Tilde{X})\in \Rb^{(seq, d_{out})}$ where the number of output is a parameter given to the layer. The KAN transformation applied is the same on each element of the sequence. On the other hand $\psi \in \Rb^{d_{out}}$ is a vector  obtained by a transformation of the signature in the GRKAN, and those weights are applied similarly at each element of the sequences. The output of the layers still has the sequence dimension, the SigKAN layers are thus stackable. Therefore, we can generalize a SigKANs network as follows:
\begin{equation}
    \begin{split}
        h_0 &= x,\\
h_j &= \text{SoftMax} (\text{GRKAN}_j(S(\tilde{h}_{j-1}))) \odot \text{KAN}_j(h_{j-1}) ,\\
y &= h_L,
    \end{split}
\end{equation}
where $ j = 1, 2, \ldots, L$. The output of the SigKANs still has the sequence dimension, and it would require either to flatten the output, only select the last outputs on this dimension, or apply an RNN that only returns the last hidden state in order to remove this dimension in the network after it.
Even if they are stackable, we have observed, in our experiments, that most of the time this is non-desirable as it doesn't yield to an increase in performance, as well as increasing the computational cost quickly. 
The size of the signature depends on the input shape raised to the power of the degree used. This, combined with the lack of GPU support for the signature computations, limits the practical viability of stacking many SigKAN layers, even though this can be beneficial when using standard fully connected layers instead of KAN layers.

\section{Learning Task}
To illustrate the benefits of our architecture, we performed two different tasks, one being the same as in \cite{genet2024tkan} and \cite{genet2024temporal} for easy comparison, the other being a task where these two tasks are more difficult to perform.
\subsection{Task definitions and dataset}
The first task we use involves predicting the notional amount traded on the market several times in advance. This is the task used in \cite{genet2024tkan} and \cite{genet2024temporal}, and is interesting because volumes exhibit multiple patterns such as seasonality and autocorrelation.The second task changes only the target value; instead of trying to estimate future volume, we aim to predict future absolute return over several time steps in advance. The tasks focus solely on the Binance exchange, so our dataset only contains data from this exchange, which has been the most important market for many years. We have also only used USDT markets, as this is the most widely used stablecoin, and all notionals are therefore intended in USDT.

\medskip

For the first task, our dataset consists of the notional amounts traded each hour on different assets:  BTC, ETH, ADA, XMR, EOS, MATIC, TRX, FTM, BNB, XLM, ENJ, CHZ, BUSD, ATOM, LINK, ETC, XRP, BCH and LTC, which are to be used to predict just one of them, BTC. For the second task, we took the absolute percentage change between closing prices only on BTC to predict BTC. For both, the data period spans 3 years from January 1, 2020 to December 31, 2022. 
\subsection{Data preprocessing}
Data preparation is an important task and can largely alters the quality of the prediction obtained. It is important to have features with the same scale between them, or to have the same target scale over time. To achieve this for the first task, we use two-stage scaling. The first step is to divide the values in the series by the moving median of the last two weeks. This moving median window is also shifted by the number of steps forward we want to predict, so as not to include foresight. This first preprocessing aims to make the series more stationary over time. The second pre-processing applied is a simple MinMaxScaling per asset, even if here the minimum of the series is 0, it is simply a matter of dividing by the maximum value. The aim is to scale the data in the 0, 1 interval to avoid an explosive effect when learning due to the power exponent. However, this pre-processing is adjusted on the training set, the adjustment being limited to finding the maximum value for each series, which is then used directly on the test set. This means that on the test set, it is possible to have data greater than 1, but as no optimization is used, this is not a problem. For the second task, as there is only one asset and volatility is a more stationary process over time than volumes, we divide the values only by the maximum value of the ream, in order to have values between 0 and 1 only. Finally, we split our data for both set into a training set and a test set, with a standard proportion of 80-20. This represents over 21,000 points in the training set and 5,000 in the test set. 
\subsection{Loss Function for Model Training}
Since we have a numerical prediction problem for both tasks, we have opted to optimize our model using the root mean square error (RMSE) as the loss function, whose formula is simple:
$$
\text{MSE} = \frac{1}{N} \sum_{i=1}^{N} \left(\hat{X}_{t+1}^{(i)} - X_{t+1}^{(i)}\right)^2,
$$
where \(N\) represents the number of samples in the dataset, \(\hat{X}_{t+1}^{(i)}\) denotes the predicted notional values of Bitcoin at time \(t+1\) for the \(i\)-th sample, and \(X_{t+1}^{(i)}\) are the corresponding true values. The first reason is that this is the most widely used and standard method in machine learning for this type of problem.
The second reason is the metric we want to use to display the results, namely the R-squared $R^2$. The $R^2$ is interesting as a metric because it not only gives information on the error but also on the error given the variance of the estimated series, which means it's much easier to tell whether the model is performing well or not. It is also a measure widely used by econometricians and practitioners for this very reason. However, minimizing the MSE is exactly the same as maximizing the \(R^2\), as its formula indicates:
$$
R^2 = 1 - \frac{\sum_{i=1}^{N} (\hat{X}_{t+1}^{(i)} - X_{t+1}^{(i)})^2}{\sum_{i=1}^{N} (X_{t+1}^{(i)} - \bar{X}_{t+1})^2}.
$$
As we can see, the upper terms of the quotient are equivalent to the sum of the squared errors, the other two components being fixed, the optimization of the mean squared error is totally similar to the optimization of $R^2$.

\subsubsection{Note on training details}
As in \cite{genet2024tkan} and \cite{genet2024temporal}, metrics are calculated directly on scaled data and not on unscaled data. There are two reasons for this: firstly, MinMax scaling has no impact on the metric since the minimum is 0 and the data interval is $[0,1]$; rescaling would simply have involved multiplying all the points, which would not have changed the R-squared. Regarding model optimization, we utilized the Adam optimizer, which is among the most popular options. Additionally, we designated 20\% of our training set as a validation set and incorporated two training callbacks. The first is an early learning stop, which interrupts training after 6 consecutive periods without improvement on the validation set, and restores the weights associated with the best loss obtained on the validation set. The second is a reduction in the plateau learning rate, which halves the learning rate after 3 consecutive periods without improvement on the validation set. Finally, the sequence length given to all models for both tasks is set at a maximum between 45 and 5 times the number of steps forward. We set this lower limit at 45, as we observed that this positively improved the performance of all models. The results for predictions at 1, 3 and 6 steps ahead are therefore different from those in the TKAN and TKAT papers.

\subsection{Task 1: Predicting Volumes}
\subsubsection{Model Architecture and Benchmarks}
As these tasks are the same as those used in the TKAN and TKAT articles, we decided to use the same benchmarks for comparison. We also compared SigKAN with a variant called SigDense, in which the KANLinear is replaced by a fully connected dense layer, and the GRKAN is replaced by a standard GRN layer.
\begin{enumerate}
    \item For the SigKAN and SigDense:
        \begin{enumerate}
            \item A variant with one SigKAN (resp. SigDense) layer with 100 units, which outputs is flatten and passed to a Dense 100 with activation 'relu' before a last linear layer. They are refered as SK-1 and SD-1 in the following tables.
            \item A variant with two consecutive SigKAN (resp. SigDense) layers with 20 units, which outputs is flatten and passed to a Dense 100 with activation 'relu' before a last linear layer. They are refered as SK-2 and SD-2 in the following tables.
        \end{enumerate}
         For all of them, we used a signature level of 2, as this is the lowest level possible to capture the interaction between inputs, while also being the least expensive in terms of computation, which quickly becomes limiting due to the non-GPU support of signature operations.
    \item For the TKAN, GRU and LSTM:
        \begin{enumerate}
    	\item An initial recurrent layer of 100 units that returns complete sequences
    	\item An intermediate recurrent layer of 100 units, which returns only the last hidden state.
    	\item A final dense layer with linear activation, with as many units as there is timesteps to predict ahead
    \end{enumerate}
    \item The TKAT uses a number of hidden units of 100 in each layer, with the exception of the embedding layer, which has only one unit per feature. The number of heads used is 4.
\end{enumerate} 
\subsubsection{Results}
Results are separated into two tables. The first table \ref{table:sig_kan} is the average $R^2$ obtained over five runs for SigKAN and SigDense. The second table \ref{table:sig_kan_std} is the standard deviation of $R^2$ obtained over these five runs. To analyze the stability of the model in relation to its initialization over several runs.
\begin{table}[H]
    \centering
    \caption{\(R^2\) Average: SigKAN and SigDense}
    \begin{tabular}{ccccc}
    \toprule
    Time & SK-1 & SD-1 & SK-2 & SD-2 \\
    \midrule
    1  & 0.36225 & 0.33055 & 0.30807 & 0.31907 \\
    3  & 0.21961 & 0.21532 & 0.20580 & 0.21066  \\
    6  & 0.16361 & 0.15544 & 0.15351 & 0.15836  \\
    9  & 0.13997 & 0.12768 & 0.12684 & 0.13338  \\
    12 & 0.12693 & 0.11628 & 0.11826 & 0.11814  \\
    15 & 0.11861 & 0.11097 & 0.11448 & 0.11065 \\
    \bottomrule
    \end{tabular}
    \label{table:sig_kan}
\end{table}
\begin{table}[H]
    \centering
    \caption{\(R^2\) Average: Benchmarks}
    \begin{tabular}{ccccc}
    \toprule
    Time & TKAT & TKAN & GRU & LSTM \\
    \midrule
    1  & 0.30519 & 0.33736 & 0.36513 & 0.35553 \\
    3  & 0.21801 & 0.21227 & 0.20067 & 0.06122  \\
    6  & 0.17955 & 0.13784 & 0.08250 & -0.22583  \\
    9  & 0.16476 & 0.09803 & 0.08716 & -0.29058  \\
    12 & 0.14908 & 0.10401 & 0.01786 & -0.47322  \\
    15 & 0.14504 & 0.09512 & 0.03342 & -0.40443 \\
    \bottomrule
    \end{tabular}
    \label{table:sig_kan_std}
\end{table}
\medskip
The average of $R^2$ shows that the proposed SigKAN model outperforms all the simple reference models, namely TKAN, GRU, and LSTM, while not reaching TKAT performance for this task over a longer period. The main advantage is that while TKAT and TKAN offer superior performance for one-step prediction, the SigKAN model outperforms all the other models. This indeed seems logical in the case of short-term predictions, given the very simple architecture of the SigKAN model compared to the Kolmogorov-Arnold time transformer. But it shows that, although it contains no recurrent calculus, it manages to outperform all the simple recurrent networks to which it is compared. The results also show that using the signature as a weighting mechanism is effective, since the SigDense and SigKAN versions outperform most recurrent models in the results. And, using KAN layers instead of dense layers shows an increase in performance, indicating that both components are interesting in the proposed layer. Finally, we observe that stacking several layers does not seem to be effective in improving the performance of the KAN version on this task, while it improves that of the SigDense version on longer predictions. Using KAN in this case allows us to reduce the depth of our model for this case.
\begin{table}[H]
    \centering
    \caption{\(R^2\) Standard Deviation: SigKAN and SigDense}
    \begin{tabular}{ccccc}
    \toprule
    Time & SK-1 & SD-1 & SK-2 & SD-2 \\
    \midrule
    1  & 0.02378 & 0.01220 & 0.01102 & 0.01476 \\
    3  & 0.01435 & 0.00342 & 0.00319 & 0.00429  \\
    6  & 0.00376 & 0.00820 & 0.00438 & 0.00659  \\
    9  & 0.00346 & 0.01074 & 0.00665 & 0.00385  \\
    12 & 0.00369 & 0.00106 & 0.00830 & 0.00215  \\
    15 & 0.00171 & 0.00442 & 0.00216 & 0.00103 \\
    \bottomrule
    \end{tabular}
\end{table}
\begin{table}[H]
    \centering
    \caption{\(R^2\) Standard Deviation: Benchmarks}
    \begin{tabular}{ccccc}
    \toprule
    Time & TKAT & TKAN & GRU & LSTM \\
    \midrule
    1  & 0.01886 & 0.00704 & 0.00833 & 0.01116 \\
    3  & 0.00906 & 0.00446 & 0.00484 & 0.08020 \\
    6  & 0.00654 & 0.01249 & 0.02363 & 0.06271 \\
    9  & 0.00896 & 0.02430 & 0.01483 & 0.05272 \\
    12 & 0.00477 & 0.00132 & 0.08638 & 0.08574 \\
    15 & 0.01014 & 0.00701 & 0.02407 & 0.09272 \\
    \bottomrule
    \end{tabular}
\end{table}
What is particularly intriguing about the SigKAN model is its ability to produce more stable results across various runs, especially when compared to models based on recurrent networks for longer prediction tasks. So, while increasing performance compared to simple recurrent models, the addition of the signature component eliminates the need for the recurrent part, which is probably one of the drivers of the greatest instability in the other models.
\begin{table}[H]
    \centering
    \caption{\(R^2\) Number of parameters: SigKAN and SigDense}
    \begin{tabular}{ccccc}
    \toprule
    Time & SK-1 & SK-2 & SD-1 & SD-2 \\
    \midrule
    1 & 957,846 & 266,311 & 558,946 & 125,791 \\
    3 & 958,048 & 266,513 & 559,148 & 125,993 \\
    6 & 958,351 & 266,816 & 559,451 & 126,296 \\
    9 & 958,654 & 267,119 & 559,754 & 126,599 \\
    12 & 1,108,972 & 297,452 & 710,072 & 156,932 \\
    15 & 1,259,290 & 327,785 & 860,390 & 187,265 \\
    \bottomrule
    \end{tabular}
\end{table}
\begin{table}[H]
    \centering
    \caption{\(R^2\) Number of parameters: Benchmarks}
    \begin{tabular}{cccccc}
    \toprule
    Time & TKAN & TKAT & GRU & MLP & LSTM \\
    \midrule
    1 & 119,396 & 1,047,865 & 97,001 & 95,801 & 128,501 \\
    3 & 119,598 & 1,057,667 & 97,203 & 96,003 & 128,703 \\
    6 & 119,901 & 1,073,870 & 97,506 & 96,306 & 129,006 \\
    9 & 120,204 & 1,091,873 & 97,809 & 96,609 & 129,309 \\
    12 & 120,507 & 1,129,676 & 98,112 & 125,412 & 129,612 \\
    15 & 120,810 & 1,178,279 & 98,415 & 154,215 & 129,915 \\
    \bottomrule
    \end{tabular}
\end{table}
We believe it is valuable to consider the number of model parameters. Simpler than the TKAT models, the SigKAN model used here displayed a much higher number of parameters. This is due to the flattening of its output, which is transmitted to the last fully connected layer, rather than to the layer itself, which has a limited number of parameters. However, the KAN layer itself has its own number of parameters which can increase sharply with the number of inputs, so it is often wise to have a smaller input size to feed the networks. For example, with 15 prediction steps, the SigKAN layers contain 513,663 weights, while the remaining 751,615 are on the intermediate dense layers connected to the flattened output. Changing the architecture to avoid having such high-dimensional inputs could be a way of radically reducing the size of the model.
\begin{figure}[H]
    \centering
    \includegraphics[width=0.7\linewidth]{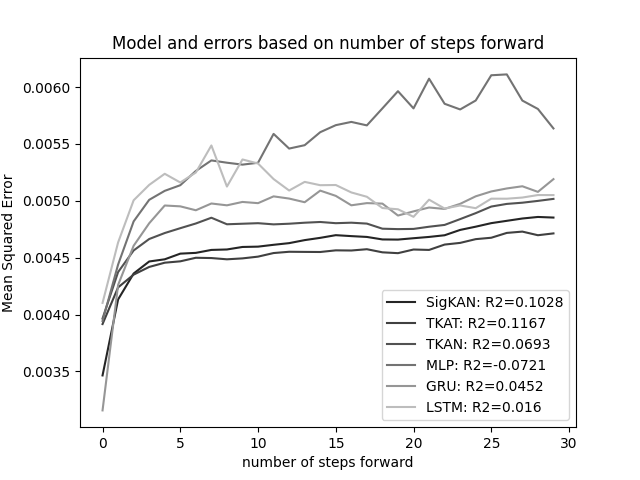}
    \caption{Model and errors based on number of steps forward} 
    \label{fig:result_graph}
\end{figure}
Finally, we trained all models once to predict the next 30 steps, and displayed for each step the $R^2$ between the predictions and the actual values. As the results above show, SigKAN lies between TKAN and TKAT in terms of performance. There's also an MLP model included here, which is the one described in the following tasks, also tested for this comparison.
\subsection{Predicting Absolute Returns: Benchmarks and Results}
\subsubsection{Model Architecture and Benchmarks}
As the task here is different, with only one feature to predict, we have chosen to keep the same TKAN, GRU and LSTM benchmarks as before, but replace TKAT with a simple MLP model. The SigKAN architecture here refers to SigKAN with 100 hidden units as in the first task, and the MLP refers to a model where sequential inputs are first flattened, then passed through 2 dense layers of 100 units with ReLU activation, before being passed to a final linear unit with the number of steps forward to predict. Volatility being a highly autocorrelated process, we also added a very simple benchmark that uses the average of previous values as a predictor of the next. The window size selected for this benchmark is the best one between 1 and 50, chosen directly from the test set, which represents the best (unrealistic) result possible with this simple method.
\subsubsection{Results}
As before, the results are obtained over 5 runs for each and displayed in an average and standard deviation table.

\begin{table}[H]
    \scriptsize
    \centering
    \caption{\(R^2\) Average: SigKAN versus Benchmarks}
    \begin{tabular}{cccccccc}
    \toprule
    Time & SK-1 & MLP & GRU & LSTM & TKAN & TKAT & Bench \\
    \midrule
    1  & 0.1670 & 0.1458 & 0.1616 & 0.1581 & 0.09063 & 0.125312 & 0.1476 \\
    3  & 0.1497 & 0.1335 & 0.1452 & 0.1488 & 0.09063 & 0.132425 & 0.1387 \\
    6  & 0.1349 & 0.1232 & 0.1360 & 0.1246 & -0.0351 & 0.132983 & 0.1293 \\
    9  & 0.1252 & 0.1172 & 0.1289 & 0.1288 & 0.00708 & 0.125162 & 0.1225 \\
    12 & 0.1164 & 0.1074 & 0.1163 & 0.1175 & -0.0716 & 0.107760 & 0.1169 \\
    15 & 0.1165 & 0.1085 & 0.1229 & 0.1169 & -0.0496 & 0.099868 & 0.1114 \\
    \bottomrule
    \end{tabular}
\end{table}
It appears that for this task, SigKAN is able to outperform all the other models on one-step-ahead predictions by a wide margin, while giving more sensible but no better predictions on longer-term predictions.The tasks also show that for these single-input tasks, the TKAN and TKAT models have much more difficulty than the other models, which may be a sign of less versatility than the SigKAN given the task. In contrast to LSTMs, which have results that are nearly the opposite of TKAN and TKAT, the latter two models achieve the weakest performance on the volume tasks but are able to considerably outperform the other models on the volatility task.
\begin{table}[H]
    \centering
    \caption{\(R^2\) Standard Deviation: SigKAN versus Benchmarks}
    \begin{tabular}{cccccccc}
    \toprule
    Time & SK-1 & MLP & GRU & LSTM & TKAN & TKAT & Bench \\
    \midrule
    1  & 0.0013 & 0.0050 & 0.0008 & 0.0003 & 0.0517 & 0.0558 & 0 \\
    3  & 0.0036 & 0.0080 & 0.0073 & 0.0012 & 0.0265 & 0.0087 & 0 \\
    6  & 0.0005 & 0.0037 & 0.0010 & 0.0020 & 0.0195 & 0.0022 & 0 \\
    9  & 0.0014 & 0.0026 & 0.0007 & 0.0022 & 0.0792 & 0.0028 & 0 \\
    12 & 0.0006 & 0.0074 & 0.0023 & 0.0006 & 0.0643 & 0.0060 & 0 \\
    15 & 0.0013 & 0.0057 & 0.0008 & 0.0001 & 0.0583 & 0.0070 & 0 \\
    \bottomrule
    \end{tabular}
\end{table}
What is also apparent is the real strength of this model: much greater stability, not only in terms of the tasks it can handle, but also in terms of stability over several calibrations, which is an important property. Finally, we also displayed the total number of parameters for each model used, and we can see that, as before, the number of trainable parameters in SigKAN is much higher than in all the others, due to the flattening of the time dimension leading to a large, dense hidden layer. Here, for 15-step forward forecasting tasks, the number of weights held by the SigKAN layers is only 114,711, while the remaining 751,615 are due to the following layers. Here again, the reduction 

\begin{table}[h]
    
    \centering
    \caption{\(R^2\) Number of parameters: SigKAN versus Benchmarks}
    \begin{tabular}{ccccccc}
    \toprule
    Time & SK-1 & TKAN & TKAT & GRU & MLP & LSTM \\
    \midrule
    1 & 563,646 & 113,906 & 485,743 & 91,601 & 14,801 & 121,301 \\
    3 & 563,848 & 114,108 & 495,545 & 91,803 & 15,003 & 121,503 \\
    6 & 564,151 & 114,411 & 511,748 & 92,106 & 15,306 & 121,806 \\
    9 & 564,454 & 114,714 & 529,751 & 92,409 & 15,609 & 122,109 \\
    12 & 714,772 & 115,017 & 567,554 & 92,712 & 17,412 & 122,412 \\
    15 & 865,090 & 115,320 & 616,157 & 93,015 & 19,215 & 122,715 \\
    \bottomrule
    \end{tabular}
\end{table}
in the number of parameters could be achieved by modifying the way the layer output is used, by not just flattening it and fully connecting all outputs to the next hidden layer.
\section{Conclusion}
The adoption of path signatures, as an improvement on traditional KANs, was presented in this paper. It was achieved through an innovative combination with Kolmogorov-Arnold networks, a technique never before realized. This new development significantly improves the ability of these networks to handle tasks involving approximation functions thanks to the use of path signatures containing rich geometric information. The results of our experimental evaluations demonstrate that SigKANs not only outperform other conventional methods for multivariate time series data, but also offer a simple robust framework for modeling complex temporal relationships, which can be applied on a large scale without too much difficulty. The fusion itself holds great promise for future research and applications: it could find its place in a variety of fields, including financial modeling or time series analysis, where such sophisticated tools are most needed. With this effort, we are going beyond mere academic interest in KANs; thanks to what we have developed here, new horizons can be opened up for machine learning, practical applications, and more generally, even other fields outside AI that could benefit from such developments.
\bibliographystyle{IEEEtran}
\bibliography{bib}

\begin{thebibliography}{10}
\providecommand{\url}[1]{#1}
\csname url@samestyle\endcsname
\providecommand{\newblock}{\relax}
\providecommand{\bibinfo}[2]{#2}
\providecommand{\BIBentrySTDinterwordspacing}{\spaceskip=0pt\relax}
\providecommand{\BIBentryALTinterwordstretchfactor}{4}
\providecommand{\BIBentryALTinterwordspacing}{\spaceskip=\fontdimen2\font plus
\BIBentryALTinterwordstretchfactor\fontdimen3\font minus \fontdimen4\font\relax}
\providecommand{\BIBforeignlanguage}[2]{{%
\expandafter\ifx\csname l@#1\endcsname\relax
\typeout{** WARNING: IEEEtran.bst: No hyphenation pattern has been}%
\typeout{** loaded for the language `#1'. Using the pattern for}%
\typeout{** the default language instead.}%
\else
\language=\csname l@#1\endcsname
\fi
#2}}
\providecommand{\BIBdecl}{\relax}
\BIBdecl

\bibitem{wei2018multivariate}
W.~W. Wei, \emph{Multivariate time series analysis and applications}.\hskip 1em plus 0.5em minus 0.4em\relax John Wiley \& Sons, 2018.

\bibitem{marcellino2006comparison}
M.~Marcellino, J.~H. Stock, and M.~W. Watson, ``A comparison of direct and iterated multistep ar methods for forecasting macroeconomic time series,'' \emph{Journal of econometrics}, vol. 135, no. 1-2, pp. 499--526, 2006.

\bibitem{tang1991time}
Z.~Tang, C.~De~Almeida, and P.~A. Fishwick, ``Time series forecasting using neural networks vs. box-jenkins methodology,'' \emph{Simulation}, vol.~57, no.~5, pp. 303--310, 1991.

\bibitem{hamilton1994state}
J.~D. Hamilton, ``State-space models,'' \emph{Handbook of econometrics}, vol.~4, pp. 3039--3080, 1994.

\bibitem{kim1994dynamic}
C.-J. Kim, ``Dynamic linear models with markov-switching,'' \emph{Journal of econometrics}, vol.~60, no. 1-2, pp. 1--22, 1994.

\bibitem{zivot2006vector}
E.~Zivot and J.~Wang, ``Vector autoregressive models for multivariate time series,'' \emph{Modeling financial time series with S-PLUS{\textregistered}}, pp. 385--429, 2006.

\bibitem{lutkepohl2013vector}
H.~L{\"u}tkepohl, ``Vector autoregressive models,'' in \emph{Handbook of research methods and applications in empirical macroeconomics}.\hskip 1em plus 0.5em minus 0.4em\relax Edward Elgar Publishing, 2013, pp. 139--164.

\bibitem{binkowski2018autoregressive}
M.~Binkowski, G.~Marti, and P.~Donnat, ``Autoregressive convolutional neural networks for asynchronous time series,'' in \emph{International Conference on Machine Learning}.\hskip 1em plus 0.5em minus 0.4em\relax PMLR, 2018, pp. 580--589.

\bibitem{ma2019novel}
K.~Ma and H.~Leung, ``A novel lstm approach for asynchronous multivariate time series prediction,'' in \emph{2019 International Joint Conference on Neural Networks (IJCNN)}.\hskip 1em plus 0.5em minus 0.4em\relax IEEE, 2019, pp. 1--7.

\bibitem{hochreiter1997long}
S.~Hochreiter and J.~Schmidhuber, ``Long short-term memory,'' \emph{Neural computation}, vol.~9, no.~8, pp. 1735--1780, 1997.

\bibitem{vaswani2017attention}
A.~Vaswani, N.~Shazeer, N.~Parmar, J.~Uszkoreit, L.~Jones, A.~N. Gomez, {\L}.~Kaiser, and I.~Polosukhin, ``Attention is all you need,'' \emph{Advances in neural information processing systems}, vol.~30, 2017.

\bibitem{liu2024kan}
Z.~Liu, Y.~Wang, S.~Vaidya, F.~Ruehle, J.~Halverson, M.~Solja{\v{c}}i{\'c}, T.~Y. Hou, and M.~Tegmark, ``Kan: Kolmogorov-arnold networks,'' \emph{arXiv preprint arXiv:2404.19756}, 2024.

\bibitem{kolmogorov1961representation}
A.~N. Kolmogorov, \emph{On the representation of continuous functions of several variables by superpositions of continuous functions of a smaller number of variables}.\hskip 1em plus 0.5em minus 0.4em\relax American Mathematical Society, 1961.

\bibitem{bozorgasl2024wav}
Z.~Bozorgasl and H.~Chen, ``Wav-kan: Wavelet kolmogorov-arnold networks,'' \emph{arXiv preprint arXiv:2405.12832}, 2024.

\bibitem{genet2024tkan}
R.~Genet and H.~Inzirillo, ``Tkan: Temporal kolmogorov-arnold networks,'' \emph{arXiv preprint arXiv:2405.07344}, 2024.

\bibitem{vaca2024kolmogorov}
C.~J. Vaca-Rubio, L.~Blanco, R.~Pereira, and M.~Caus, ``Kolmogorov-arnold networks (kans) for time series analysis,'' \emph{arXiv preprint arXiv:2405.08790}, 2024.

\bibitem{genet2024temporal}
R.~Genet and H.~Inzirillo, ``A temporal kolmogorov-arnold transformer for time series forecasting,'' \emph{arXiv preprint arXiv:2406.02486}, 2024.

\bibitem{chen1958integration}
K.-T. Chen, ``Integration of paths--a faithful representation of paths by noncommutative formal power series,'' \emph{Transactions of the American Mathematical Society}, vol.~89, no.~2, pp. 395--407, 1958.

\bibitem{lyons2014rough}
T.~Lyons, ``Rough paths, signatures and the modelling of functions on streams,'' \emph{arXiv preprint arXiv:1405.4537}, 2014.

\bibitem{lyons1998differential}
T.~J. Lyons, ``Differential equations driven by rough signals,'' \emph{Revista Matem{\'a}tica Iberoamericana}, vol.~14, no.~2, pp. 215--310, 1998.

\bibitem{chevyrev2016primer}
I.~Chevyrev and A.~Kormilitzin, ``A primer on the signature method in machine learning,'' \emph{arXiv preprint arXiv:1603.03788}, 2016.

\bibitem{yang2022developing}
W.~Yang, T.~Lyons, H.~Ni, C.~Schmid, and L.~Jin, ``Developing the path signature methodology and its application to landmark-based human action recognition,'' in \emph{Stochastic Analysis, Filtering, and Stochastic Optimization: A Commemorative Volume to Honor Mark HA Davis's Contributions}.\hskip 1em plus 0.5em minus 0.4em\relax Springer, 2022, pp. 431--464.

\bibitem{li2017lpsnet}
C.~Li, X.~Zhang, and L.~Jin, ``Lpsnet: a novel log path signature feature based hand gesture recognition framework,'' in \emph{Proceedings of the IEEE International Conference on Computer Vision Workshops}, 2017, pp. 631--639.

\bibitem{gyurko2013extracting}
L.~G. Gyurk{\'o}, T.~Lyons, M.~Kontkowski, and J.~Field, ``Extracting information from the signature of a financial data stream,'' \emph{arXiv preprint arXiv:1307.7244}, 2013.

\bibitem{dyer2021deep}
J.~Dyer, P.~W. Cannon, and S.~M. Schmon, ``Deep signature statistics for likelihood-free time-series models,'' in \emph{ICML Workshop on Invertible Neural Networks, Normalizing Flows, and Explicit Likelihood Models}, 2021.

\bibitem{perez2018signature}
I.~Perez~Arribas, G.~M. Goodwin, J.~R. Geddes, T.~Lyons, and K.~E. Saunders, ``A signature-based machine learning model for distinguishing bipolar disorder and borderline personality disorder,'' \emph{Translational psychiatry}, vol.~8, no.~1, p. 274, 2018.

\bibitem{fermanian2021embedding}
A.~Fermanian, ``Embedding and learning with signatures,'' \emph{Computational Statistics \& Data Analysis}, vol. 157, p. 107148, 2021.

\bibitem{hornik1989multilayer}
K.~Hornik, M.~Stinchcombe, and H.~White, ``Multilayer feedforward networks are universal approximators,'' \emph{Neural networks}, vol.~2, no.~5, pp. 359--366, 1989.

\bibitem{rosenblatt1958perceptron}
F.~Rosenblatt, ``{The perceptron: A probabilistic model for information storage and organization in the brain.}'' \emph{Psychological Review}, vol.~65, pp. 386--408, 1958.

\bibitem{cybenko1989approximation}
G.~Cybenko, ``Approximation by superpositions of a sigmoidal function,'' \emph{Mathematics of control, signals and systems}, vol.~2, no.~4, pp. 303--314, 1989.

\bibitem{lim2021temporal}
B.~Lim, S.~{\"O}. Ar{\i}k, N.~Loeff, and T.~Pfister, ``Temporal fusion transformers for interpretable multi-horizon time series forecasting,'' \emph{International Journal of Forecasting}, vol.~37, no.~4, pp. 1748--1764, 2021.

\bibitem{elfwing2018sigmoid}
S.~Elfwing, E.~Uchibe, and K.~Doya, ``Sigmoid-weighted linear units for neural network function approximation in reinforcement learning,'' \emph{Neural networks}, vol. 107, pp. 3--11, 2018.

\bibitem{clevert2015fast}
D.-A. Clevert, T.~Unterthiner, and S.~Hochreiter, ``Fast and accurate deep network learning by exponential linear units (elus),'' \emph{arXiv preprint arXiv:1511.07289}, 2015.

\bibitem{ba2016layer}
J.~L. Ba, J.~R. Kiros, and G.~E. Hinton, ``Layer normalization,'' 2016.

\bibitem{dauphin2017language}
Y.~N. Dauphin, A.~Fan, M.~Auli, and D.~Grangier, ``Language modeling with gated convolutional networks,'' in \emph{International conference on machine learning}.\hskip 1em plus 0.5em minus 0.4em\relax PMLR, 2017, pp. 933--941.

\end{thebibliography}
\end{document}